# Intuitive Human-Robot Interaction: Development and Evaluation of a Gesture-Based User Interface for Object Selection

Bijan KAVOUSIAN, Oliver PETROVIC, Werner HERFS

*Laboratory for Machine Tools and Production Engineering WZL of RWTH Aachen Steinbachstraße 25, D-52072 Aachen*

**Abstract:** Gestures are a natural form of communication between humans and should also be leveraged for human-robot interaction. This work presents a gesture-based user interface for object selection using pointing and click gestures. An experiment with 20 participants evaluates accuracy and selection time, demonstrating the potential for efficient collaboration.

**Keywords:** Computer Vision, Gesture Control, Intuitive Interaction, Human-Robot Collaboration, Robotics

## 1. Introduction

In industrial automation, the use of robots is becoming increasingly important. To simplify deployment, no-code programming approaches are gaining attention as an alternative to classical programming methods. Effective communication between humans and robots is therefore essential. In addition to speech-based interaction, gestures play a central role in human-robot interaction and are widely studied (Wang et al. 2022).

Pointing with a finger is a natural and effective way for humans to identify objects. This is particularly beneficial in industrial environments, where objects often lack intuitive names or appear multiple times. This intuitive form of communication is transferred to human-robot interaction in this work. The contributions are: (1) A calibration-free, real-time capable system for object selection, (2) a unified pipeline combining modern segmentation and hand pose estimation methods, and (3) An experimental study with 20 participants analyzing the influence of pointing strategies and visual feedback

## 2. Related Work

Previous work has already investigated pointing gestures for object selection in human–robot interaction. Quintero et al. (2013) implemented a system for the visual recognition of pointing gestures based on estimating a pointing direction using lines between the hand and the head or elbow. In a later study, Quintero et al. (2015) showed that such pointing gestures can be used both for communication from the human to the robot and as robot-generated feedback. However, object selection success rates range only between 51% and 69% (Quintero et al. 2013), and without modern object detection, the distinction between individual objects remained imprecise.

Further approaches employ pointing gestures in specific assistive contexts. Jevtic et al. (2019) use pointing gestures to support the process of putting on shoes. However, this requires predefined shoe positions, and both the gestures and the scene must be calibrated, which strongly limits flexibility. Similarly, Moh et al. (2019) use gesture-based interaction in a household context, again without general object recognition for arbitrary scenes.

Recent surveys show that gestures remain a central research area in HRI, particularly for natural interaction and intuitive object referencing (Wang et al. 2022). Modern systems such as Ekrekli et al. (2023) combine gestures and speech for multimodal interaction. However, object selection is not based on actual finger pointing but instead on selecting the object closest to the wrist. This does not correspond to the natural human interpretation of pointing and can lead to misassignments, especially for objects that are further away.

Overall, these works show that pointing gestures can in principle be used as an interaction modality, but so far mostly without general, instance-based object recognition, true interpretation of finger pointing, and integration into a unified real-time system. Flexible object selection in varying scenes using only a single RGB-D camera has therefore not yet been addressed.

### 3. Technological Concept and Implementation

To implement the system, hand tracking and object detection are combined with a user interface into an overall system. This enables users to select objects using pointing gestures. The system uses RGB-D data from an Intel RealSense D435i camera. After objects are segmented during object detection and their spatial centroids are determined using depth information, a pointing line is constructed through the hand during hand tracking (see Figure 2), and the object closest to this pointing line is preselected. As soon as a selection gesture is performed with the other hand, the object is selected. Figure 1 provides an overview of the overall system.

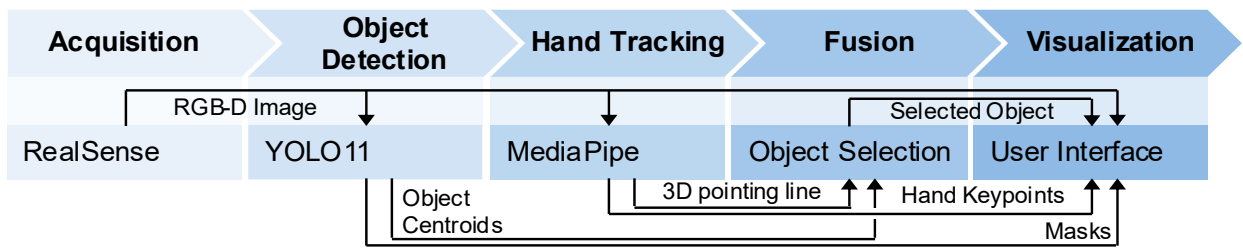

*Figure 1: Overview of the overall system*

3.1 Object Detection

Starting from the camera image, objects are first segmented in the two-dimensional image. Various segmentation algorithms were tested for this purpose. SAM2 (Ravi et al. 2024) is a foundation model for segmenting general images but is not fast enough for real-time applications. MobileSAM (Zhang et al. 2023) is faster, but its accuracy is not sufficient for the application. Therefore, the pretrained YOLO11-seg (Khanam und Hussain 2024) is used, which can segment a wide variety of objects without additional training. After segmentation, a spatial centroid is determined for each object using depth data, which is then used for object selection.

*3.2 Hand Tracking*

For hand tracking, the implementation from Google's MediaPipe framework (Lugaresi et al. 2019) is used, which provides a pretrained, real-time capable model for detecting hand keypoints. The keypoints are shown in Figure 2 (real image: Figure 3) and are used for two purposes: the points of one hand are used to construct a pointing line, while the points of the other hand are used to detect the selection gesture.

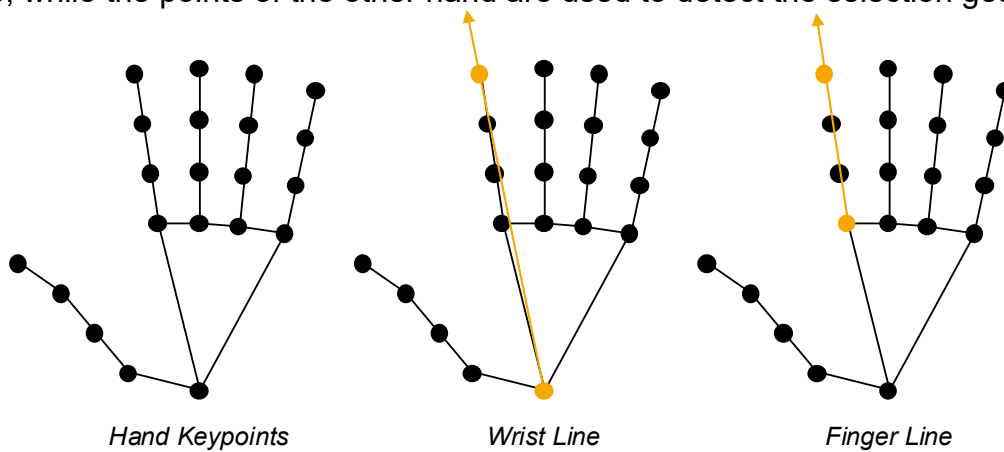

*Hand Keypoints*  *Wrist Line*  *Finger Line*

***Figure 2:*** *Detected hand keypoints and evaluated pointing lines*

Two different pointing lines were tested: the line between the first joint and the tip of the index finger ("finger line") corresponds to the human understanding of pointing, while the line from the wrist to the tip of the index finger ("wrist line") follows a similar direction but is expected to yield more stable results due to the greater distance between supporting points. Both variants are compared later in the experiments. The two-dimensional line is extended into a three-dimensional line in space using depth information (cf. Quintero et al. 2013).

To perform object selection, a click gesture is introduced. This click gesture is established in VR systems such as the Apple Vision Pro (Apple 2025) and is intuitively applicable. For this purpose, the distance between the thumb tip and the index finger tip of the second hand is computed using the keypoints. As soon as it falls below a threshold, the object is selected.

## 4. Experimental Setup and Procedure

To evaluate the method, experiments were conducted with 20 participants who were not familiar with the system prior to the study. In particular, it was investigated which pointing line provides better selection accuracy and what influence the visual user interface has. All participants consented to take part and were able to withdraw from the experiment at any time without consequences.

The experimental setup is shown in Figure 3. Six objects were placed on a table, behind which a screen displaying the user interface was positioned. The camera image, the segmentation masks of the objects, and the detected hand keypoints are displayed there. As soon as an object can be selected, it is highlighted in green in the visualization.

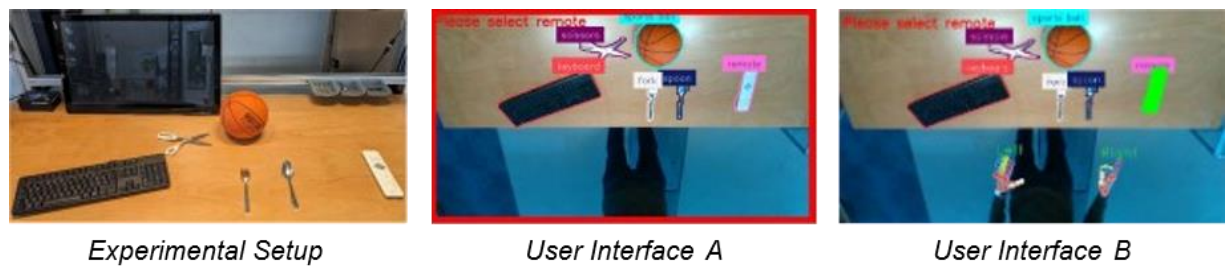

*Figure 3: Experimental setup and user interface. A: The objects are segmented; a red frame indicates that the hands are not in the image. B: The hands are detected and keypoints are visualized; the pre-selected object is highlighted.*

Participants were instructed via the screen to select a specific object. This ensures that their intention is known and can be compared with the system output. The target object is displayed on the screen (rather than being spoken aloud) to increase comparability and avoid introducing additional temporal uncertainty. In each trial, every object is selected twice in random order.

To compare the pointing lines and to demonstrate the benefit of the visual interface, a 2×2 factorial within-subject design is used. Each participant performs four evaluated conditions: with and without visual user interface, and with finger line and wrist line. The trials are conducted in random order. Additionally, participants perform a training trial with visual feedback before the evaluated trials to reduce the influence of learning effects. After completing all trials, participants fill out a questionnaire with Likert-scale questions and a field for free-text comments.

## 5. Results

The analysis is based on the selection accuracies determined per participant in the four experimental conditions. Table 1 shows the means and standard deviations for the combinations of pointing line (finger vs. wrist) and visual feedback (on vs. off).

*Table 1: Means and standard deviations for the different conditions*

| Pointing line | Visualization | Mean | Standard deviation |
|---|---|---|---|
| Finger | On | 93.3 % | 7.3 % |
| Wrist | On | 91.5 % | 8.9 % |
| Finger | Off | 90.8 % | 8.1 % |
| Wrist | Off | 82.1 % | 17.6 % |

For the analysis, a two-factor repeated-measures ANOVA was conducted. The factors were pointing line (finger vs. wrist) and visual feedback (on vs. off). The ANOVA shows a significant main effect of visual feedback on selection accuracy ($p = 0.037$). With visual feedback, objects were selected more reliably. The main effect of pointing line is not significant ($p = 0.076$), but is close to the significance level and indicates a possible effect with a larger sample size. The interaction between both factors is not significant ($p = 0.162$).

For selection time, the ANOVA showed no significant effects. The mean selection time was approximately three to four seconds across all conditions. This value includes both reading and understanding the instruction, as well as the subsequent reaction and the actual selection using the system.

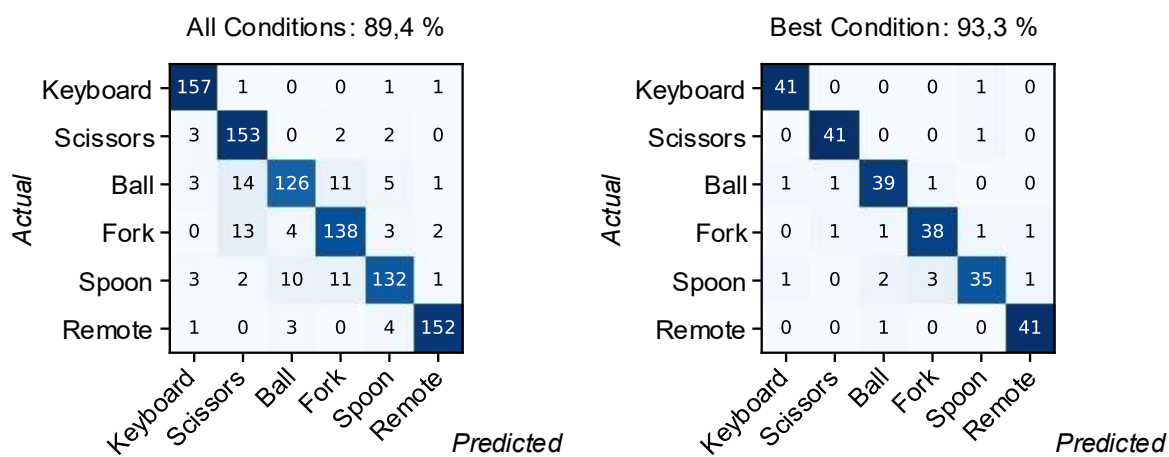

*Figure 4: Confusion matrix across all conditions as well as for the best-performing condition with visual interface and finger line*

Figure 4 first shows a confusion matrix aggregated over all trials. The overall accuracy in this case is 89.4%. It can be observed that objects located close to each other are more frequently confused than objects that are further apart. In the best-performing condition, with visual feedback enabled and using the finger line, proportionally fewer errors occur, and the accuracy reaches 93.3%.

The questionnaire provides additional subjective assessments from the participants. The selection process was predominantly described as understandable, and both the pointing gesture and the click gesture were mostly rated as understandable to very understandable. Visual feedback was often perceived as helpful, while some participants simultaneously reported that the interaction felt more natural without feedback. Due to the small sample size, these responses should be interpreted as a qualitative complement to the objective results.

## 6. Conclusion and Future Work

This work investigates a gesture-based user interface for object selection in human–robot interaction and shows that the developed system can reliably identify and select objects. Visual feedback proved to be a key factor for achieving high selection accuracy, while the finger line showed slight advantages over the wrist-based pointing line. The subjective assessments of the participants generally confirm the understandability and transparency of the interaction concept.

Overall, the results demonstrate that simple pointing and click gestures can provide a robust and accessible form of interaction and are well suited for intuitive selection in a shared workspace. Especially for objects that are located close to each other, this approach offers advantages for clear object identification compared to others, such as the method proposed by Ekrekli et al.

In future work, the system can be further improved through more powerful segmentation and recognition methods and integrated into a robotic system, enabling gesture-based selection to seamlessly transition into concrete robotic actions.

**Acknowledgements:** The studies described in this paper were conducted as part of the research and development project "AKzentE4.0". The project is funded by the German Federal Ministry of Research, Technology and Space (BMFTR) under the funding measure "Future of Work: Regional Competence Centers of Labor Research. Designing New Forms of Work through Artificial Intelligence" within the program "Innovations for Tomorrow's Production, Services and Work" (funding code: 02L19C400) and supervised by the Project Management Agency Karlsruhe (PTKA).

# Intuitive Mensch-Roboter-Interaktion: Entwicklung und Evaluation eines gestenbasierten User Interfaces für die Objektauswahl

Bijan KAVOUSIAN, Oliver PETROVIC, Werner HERFS

*Werkzeugmaschinenlabor WZL der RWTH Aachen
Steinbachstraße 25, D-52072 Aachen*

**Kurzfassung:** Gesten sind eine natürliche Form der Kommunikation zwischen Menschen und sollen auch für die Mensch-Roboter-Interaktion nutzbar gemacht werden. Die Arbeit präsentiert ein gestenbasiertes User Interface zur Objektauswahl durch Fingerzeigen und Klickgesten. Ein Experiment mit 20 Teilnehmenden untersucht Genauigkeit und Auswahlzeit und zeigt das Potenzial für eine effiziente Zusammenarbeit.

**Schlüsselwörter:** Computer Vision, Gestensteuerung, Intuitive Interaktion, Mensch-Roboter-Kollaboration, Robotik

## 1. Einleitung

In der industriellen Automatisierung wird der Einsatz von Robotern immer relevanter. Zur vereinfachten Einrichtung rückt dabei anstelle klassischer Programmiermethoden zunehmend die No-Code-Programmierung in den Fokus. Dafür ist eine effektive Kommunikation zwischen Mensch und Roboter unerlässlich. Neben sprachlichen Interaktionsformen spielen Gesten auch in der Mensch-Roboter-Interaktion eine zentrale Rolle und werden intensiv erforscht (Wang et al. 2022). Besonders das Zeigen mit dem Finger dient zwischen Menschen der eindeutigen Identifikation von Objekten, was im industriellen Kontext vorteilhaft ist, da Objekte oft keine intuitiven Namen haben oder mehrfach vorkommen.

Diese Form der intuitiven Kommunikation soll daher auf die Interaktion zwischen Mensch und Roboter übertragen werden. Die vorliegende Arbeit trägt hierzu bei, indem sie (1) ein kalibrierungsfreies, echtzeitfähiges System zur Objektauswahl entwickelt, (2) moderne Verfahren zur Segmentierung und Handposen-Erkennung in einer durchgängigen Pipeline kombiniert und (3) in einem Versuch mit 20 Personen den Einfluss der Zeigelinie sowie der visuellen Rückmeldung untersucht.

## 2. Stand der Technik

Vorangegangene Arbeiten untersuchen bereits Zeigegesten zur Objektauswahl in der Mensch-Roboter-Interaktion. Quintero et al. (2013) implementieren ein System zur visuellen Erkennung von Zeigegesten, das auf der Schätzung einer Zeigerichtung durch Linien zwischen Hand und Kopf bzw. Ellbogen basiert. In einer späteren Arbeit zeigen Quintero et al. (2015), dass solche Zeigegesten sowohl zur Kommunikation vom Menschen zum Roboter als auch als robotergeneriertes Feedback genutzt werden können. Die Erfolgsquoten der Objektauswahl liegen jedoch lediglich zwischen 51 % und 69 % (Quintero et al. 2013), und ohne moderne Objekterkennung blieb die Abgrenzung einzelner Objekte ungenau.

Weitere Ansätze setzen Zeigegesten in spezifischen Assistenzkontexten ein. Jevtic et al. (2019) verwenden Zeigegesten zur Unterstützung beim Anziehen von Schuhen. Hierfür müssen jedoch feste Schuhpositionen definiert und sowohl die Gesten als auch die Szene kalibriert werden, sodass die Flexibilität stark eingeschränkt ist. Ähnlich verwenden Moh et al. (2019) gestenbasierte Interaktion in einem Haushaltskontext, jedoch wieder ohne allgemeine Objekterkennung für beliebige Szenen.

Aktuelle Übersichtsarbeiten zeigen, dass Gesten weiterhin ein zentrales Forschungsfeld der HRI bilden, insbesondere für natürliche Interaktion und intuitive Objektreferenzierung (Wang et al. 2022). Moderne Systeme wie Ekrekli et al. (2023) kombinieren Gesten und Sprache für multimodale Interaktion. Die Objektauswahl erfolgt dort jedoch nicht über echtes Fingerzeigen, sondern über die Wahl des zum Handgelenk nächstgelegenen Objekts. Das entspricht nicht der natürlichen menschlichen Interpretation eines Zeigens, wodurch es besonders bei weiter entfernten Objekten zu Fehlzuordnungen kommen kann.

Insgesamt zeigen diese Arbeiten, dass Zeigegesten grundsätzlich als Interaktionsmittel genutzt werden können, jedoch bisher meist ohne allgemeine, instanzbasierte Objekterkennung, echte Fingerzeiginterpretation und Integration in ein durchgängiges Echtzeitsystem. Eine flexible Objektauswahl in variablen Szenen mit nur einer RGB-D-Kamera wird damit bislang nicht adressiert.

## 3. Technologisches Konzept und Umsetzung

Um das System umzusetzen, werden Handtracking und Objekterkennung mit einer Nutzeroberfläche zu einem Gesamtsystem zusammengesetzt. Damit können die Nutzenden Objekte mit Zeigegesten auswählen. Das System verwendet dabei RGB-D-Daten aus einer Kamera von Intel des Typs Realsense D435i. Nachdem bei der Objekterkennung die Objekte segmentiert und mithilfe der Tiefeninformationen die räumlichen Mittelpunkte bestimmt wurden, wird beim Handtracking eine Zeigelinie durch die Hand (siehe Abb. 2) gelegt und das nächste Objekt zur Zeigelinie vorausgewählt. Sobald mit der anderen Hand eine Auswahlgeste durchgeführt wird, wird das Objekt ausgewählt. Abbildung 1 gibt einen Überblick über das Gesamtsystem.

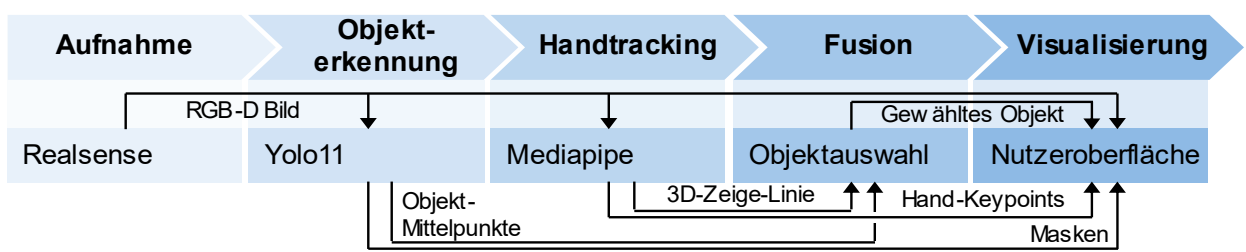

***Abbildung 1:*** *Überblick über das Gesamtsystem.*

### 3.1 Objekterkennung

Ausgehend vom Kamerabild werden zunächst die Objekte im zweidimensionalen Bild segmentiert. Dafür wurden verschiedene Segmentierungsalgorithmen getestet. SAM2 (Ravi et al. 2024) ist ein Foundation Model zur Segmentierung allgemeiner Bilder, ist allerdings für die Echtzeitanwendung nicht schnell genug. MobileSam (Zhang et al. 2023) ist schneller, die Genauigkeit reicht jedoch nicht für die Anwendung aus. Daher wird das vortrainierte YOLOv11seg (Khanam und Hussain 2024) verwendet, das eine Vielzahl von Objekten ohne zusätzliches Training segmentieren kann. Nach der Segmentierung wird für jedes Objekt mithilfe der Tiefendaten ein Raummittelpunkt bestimmt, der für die Objektauswahl verwendet wird.

*3.2 Handtracking*

Für das Handtracking wird die Implementierung aus Googles Mediapipe-Framework (Lugaresi et al. 2019) verwendet, das ein vortrainiertes, echtzeitfähiges Modell zur Erkennung von Hand-Keypoints bereitstellt. Die Keypoints sind in Abbildung 2 dargestellt (reales Bild: Abbildung 3) und werden für zwei Zwecke verwendet: Die Punkte einer Hand werden genutzt, um eine Zeigelinie zu konstruieren. Mit den Punkten der anderen Hand wird die Auswahlgeste erkannt.

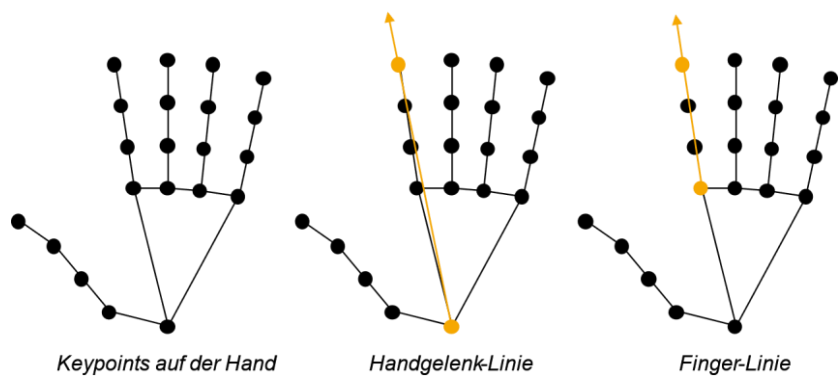

***Abbildung 2:*** *Erkannte Keypoints auf der Hand sowie getestete Zeige-Linien.*

Für die Zeigelinien wurden zwei verschiedene Linien getestet: Die Linie zwischen dem ersten Gelenk und der Spitze des Zeigefingers („Finger-Linie") entspricht dem Zeigeverständnis des Menschen, während die Linie vom Handgelenk zur Fingerspitze des Zeigefingers („Handgelenk-Linie") ähnlich verläuft, aber durch die weiter auseinanderliegenden Stützpunkte stabilere Ergebnisse erwarten lässt. Beide Varianten werden später in den Versuchen verglichen. Die zweidimensionale Linie wird mithilfe der Tiefeninformationen zu einer dreidimensionalen Linie im Raum ergänzt (vgl. Quintero et al. 2013).

Um die Objektauswahl durchzuführen, wird eine Klickgeste eingeführt. Diese Klickgeste ist bei VR-Systemen wie der Apple Vision Pro etabliert (Apple 2025) und intuitiv anwendbar. Hierzu wird mithilfe der Keypoints der zweiten Hand der Abstand zwischen Daumen- und Zeigefingerspitze ermittelt. Sobald er unter einen Schwellwert fällt, wird das Objekt ausgewählt.

## 4. Versuchsaufbau und Durchführung

Zur Bewertung der Methode wurden Experimente mit 20 Teilnehmenden durchgeführt, die das System vor den Versuchen nicht kannten. Dabei wurde insbesondere untersucht, welche Zeigelinie die bessere Auswahlgenauigkeit ermöglicht und welchen Einfluss die visuelle Benutzeroberfläche hat. Alle Teilnehmenden stimmten der Teilnahme zu und konnten das Experiment jederzeit ohne Konsequenzen abbrechen.

Der Versuchsaufbau ist in Abbildung dargestellt. Auf einem Tisch wurden sechs Objekte platziert, hinter denen sich ein Bildschirm mit der Benutzeroberfläche befindet.

Dort werden das Kamerabild, die Segmentierungsmasken der Objekte sowie die erkannten Handpunkte angezeigt. Sobald ein Objekt ausgewählt werden kann, wird es in der Darstellung grün hervorgehoben.

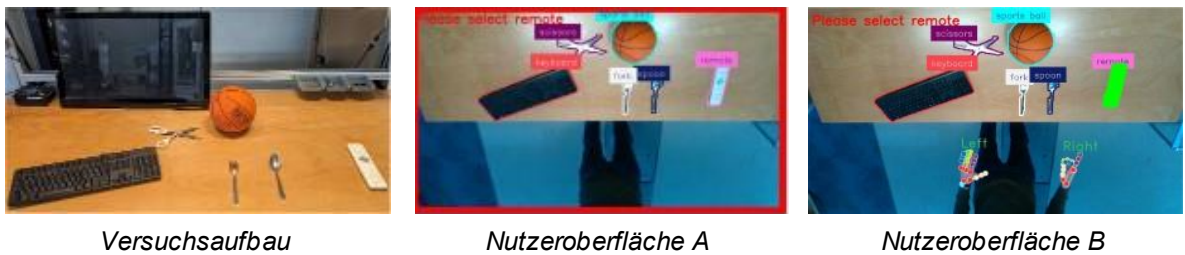

*Versuchsaufbau*     *Nutzeroberfläche A*     *Nutzeroberfläche B*

***Abbildung 3:*** *Versuchsaufbau und Nutzeroberfläche. A: Die Objekte werden segmentiert, ein roter Rahmen zeigt an, dass die Hände sich nicht im Bild befinden. B: Die Hände werden erkannt und Keypoints eingezeichnet; das vorausgewählte Objekt wird hervorgehoben.*

Die Versuchspersonen erhalten auf dem Bildschirm die Anweisung, ein spezifisches Objekt auszuwählen. Dadurch ist ihre Absicht bekannt und kann mit der Systemausgabe verglichen werden. Das auszuwählende Objekt wird auf dem Bildschirm angezeigt (und nicht etwa vorgelesen), um die Vergleichbarkeit zu erhöhen und keine zusätzliche zeitliche Unsicherheit einzuführen. Pro Versuchsdurchlauf wird jedes Objekt zweimal in zufälliger Reihenfolge ausgewählt.

Um die Zeigelinien zu vergleichen sowie den Nutzen der visuellen Oberfläche nachzuweisen, wird ein 2×2-faktorieller Versuchsplan mit Messwiederholung verwendet. Dabei führt jede Versuchsperson vier ausgewertete Varianten des Versuchs durch: Mit bzw. ohne visuelle Nutzeroberfläche und mit Finger-Linie und Handgelenk-Linie. Die Versuche finden in einer zufälligen Reihenfolge statt. Zusätzlich führen die Versuchspersonen vor den ausgewerteten Versuchen einen Übungsversuch mit visuellem Feedback durch, um den Einfluss von Lerneffekten auf die Auswertung zu verringern. Nach allen Versuchen füllen die Versuchspersonen einen Fragebogen mit Fragen auf einer Likert-Skala sowie einem Feld für freie Kommentare aus.

## 5. Ergebnisse

Die Auswertung basiert auf den pro Versuchsperson ermittelten Auswahlgenauigkeiten in den vier experimentellen Bedingungen. Tabelle 1 zeigt die Mittelwerte und Standardabweichungen für die Kombinationen aus Zeigelinie (Finger vs. Handgelenk) und visueller Rückmeldung (an vs. aus).

***Tabelle 1:*** *Mittelwerte und Standardabweichungen für die verschiedenen Versuche.*

| **Zeige-Linie** | **Visualisierung** | **Mittelwert** | **Standardabweichung** |
|---|---|---|---|
| Finger | An | 93,3 % | 7,3 % |
| Handgelenk | An | 91,5 % | 8,9 % |
| Finger | Aus | 90,8 % | 8,1 % |
| Handgelenk | Aus | 82,1 % | 17,6 % |

Für die Analyse wurde eine zweifaktorielle ANOVA mit Messwiederholung durchgeführt. Die Faktoren waren Zeigelinie (Finger vs. Handgelenk) und visuelle Rückmeldung (an vs. aus). Die ANOVA zeigt einen signifikanten Haupteffekt der visuellen

Rückmeldung auf die Auswahlgenauigkeit ($p = 0.037$). Mit visueller Rückmeldung wurden Objekte zuverlässiger ausgewählt. Der Haupteffekt der Zeigelinie ist nicht signifikant ($p = 0.076$), liegt jedoch nahe am Signifikanzniveau und deutet auf einen möglichen Einfluss bei größerer Stichprobe hin. Die Interaktion zwischen beiden Faktoren ist nicht signifikant ($p = 0.162$).

Für die Auswahlzeit ergab die ANOVA keine signifikanten Effekte. Die mittlere Auswahlzeit lag in allen Bedingungen im Bereich von etwa drei bis vier Sekunden. Dieser Wert umfasst sowohl das Lesen und Verstehen der Anweisung als auch die anschließende Reaktion und die eigentliche Auswahl mit dem System.

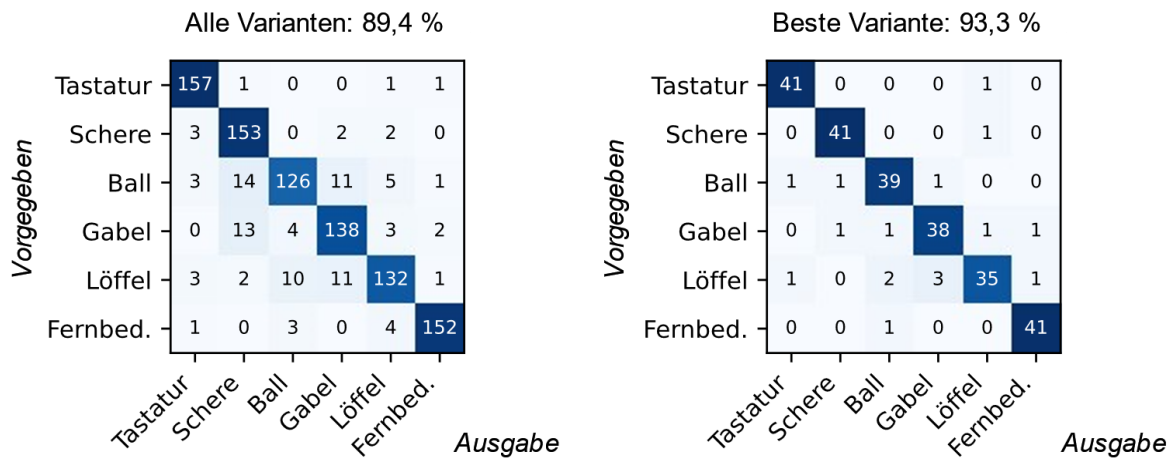

*Abbildung 4:* Confusion Matrix: Über alle Varianten hinweg sowie für die am besten funktionierende Variante mit visueller Oberfläche und Finger-Linie.

In Abbildung 4 ist zunächst eine Confusion Matrix über alle Versuche hinweg abgebildet. Die gesamte Genauigkeit ist in diesem Fall 89,4 %. Es fällt auf, dass nah aneinanderliegende Objekte häufiger verwechselt werden als weiter voneinander entfernte Objekte. In der besten Variante mit angeschalteter Visualisierung und Finger-Linie kommt es proportional zu weniger Fehlern, die Genauigkeit liegt bei 93,3 %.

Die Befragung liefert zusätzliche subjektive Einschätzungen der Teilnehmenden. Der Auswahlprozess wurde überwiegend als nachvollziehbar beschrieben, und sowohl Zeigegeste als auch Klickgeste wurden meist als verständlich bis sehr verständlich eingestuft. Die visuelle Rückmeldung wurde häufig als hilfreich wahrgenommen, während einige Teilnehmende gleichzeitig angaben, dass sich die Interaktion ohne Rückmeldung natürlicher anfühle. Aufgrund der kleinen Stichprobe sind die Angaben als qualitative Ergänzung zu den objektiven Ergebnissen zu verstehen.

## 6. Fazit und Ausblick

Die Arbeit untersucht ein gestenbasiertes User Interface zur Objektauswahl in der Mensch-Roboter-Interaktion und zeigt, dass das entwickelte System Objekte zuverlässig identifizieren und auswählen kann. Die visuelle Rückmeldung erwies sich dabei als zentraler Faktor für eine hohe Auswahlgenauigkeit, während die Finger-Linie leichte Vorteile gegenüber der Zeigelinie über das Handgelenk zeigte. Die subjektiven Einschätzungen der Teilnehmenden bestätigen grundsätzlich die Verständlichkeit und Nachvollziehbarkeit des Interaktionskonzepts. Insgesamt verdeutlichen die Ergebnisse, dass einfache Zeige- und Klickgesten eine robuste und zugängliche Form der Interaktion bieten können und sich gut für eine intuitive Auswahl im gemeinsamen Arbeitsraum eignen. Gerade bei nah aneinander liegenden Objekten bietet dieser Ansatz im Vergleich zu anderen wie z. B. von Ekrekli et al. Vorteile für die eindeutige Identifikation von Objekten. Zukünftig kann das System durch leistungsfähigere Segmentierungs- und Erkennungsverfahren weiter verbessert und in ein Robotersystem eingebunden werden, sodass die gestenbasierte Auswahl nahtlos in konkrete robotische Handlungen übergeht.

## 7. Literatur